\documentclass[10pt,twocolumn,letterpaper]{article}

\usepackage{iccv}
\usepackage{times}
\usepackage{graphicx}
\usepackage[dvipsnames]{xcolor}
\usepackage{amsmath}
\usepackage{amssymb}
\usepackage{pifont}
\usepackage{tabularx}
\usepackage{makecell}
\usepackage{caption}
\usepackage{subcaption}
\usepackage{bbold}
\usepackage{booktabs}
\usepackage{enumitem}

\usepackage[pagebackref=true,breaklinks=true,letterpaper=true,colorlinks,bookmarks=false]{hyperref}

\iccvfinalcopy 

\def\iccvPaperID{48} 

\ificcvfinal\fi

\DeclareMathOperator*{\argmax}{arg\,max} 

\newcommand{\xmark}{\ding{55}}

\newcolumntype{P}[1]{>{\centering\arraybackslash}p{#1}}

\newcommand{\myparagraph}[1]{\vspace{4pt}\noindent\textbf{#1}}

\usepackage[capitalize]{cleveref}
\crefname{section}{Sec.}{Secs.}
\crefname{table}{Tab.}{Tabs.}
\crefname{figure}{Fig.}{Figs.}

\newcommand{\myalgname}{CoRTe}
\newcommand{\setting}{{Source-Free Black-Box Unsupervised Domain Adaptation for Semantic Segmentation}}

\begin{document}

\title{\vspace{-5pt} Cross-Domain Transfer Learning with CoRTe:  Consistent and Reliable Transfer from Black-Box to Lightweight Segmentation Model}

\author{
Claudia Cuttano
\and
Antonio Tavera
\and
Fabio Cermelli \and 
Giuseppe Averta
\and
Barbara Caputo\\
{\tt\small name.surname@polito.it}\\
Politecnico di Torino, Corso Duca degli Abruzzi, 24 — 10129 Torino, ITALIA\\
}

\maketitle
\ificcvfinal\pagestyle{empty}\fi

\begin{abstract}
Many practical applications require training of semantic segmentation models on unlabelled datasets and their execution on low-resource hardware. 
Distillation from a trained source model may represent a solution for the first but does not account for the different distribution of the training data. Unsupervised domain adaptation (UDA) techniques claim to solve the domain shift, but in most cases assume the availability of the source data or an accessible white-box source model, which in practical applications are often unavailable for commercial and/or safety reasons.
In this paper, we investigate a more challenging setting in which a lightweight model has to be trained on a target unlabelled dataset for semantic segmentation, under the assumption that we have access only to black-box source model predictions.
Our method, named {\myalgname}, consists of (i) a pseudo-labelling function that extracts reliable knowledge from the black-box source model using its relative confidence, (ii) a pseudo label refinement method to retain and enhance the novel information learned by the student model on the target data, and (iii) a consistent training of the model using the extracted pseudo labels.
We benchmark {\myalgname} on two synthetic-to-real settings, demonstrating remarkable results when using black-box models to transfer knowledge on lightweight models for a target data distribution. 
\end{abstract}

\begin{figure}
\centering
  \includegraphics[width=\linewidth]
  {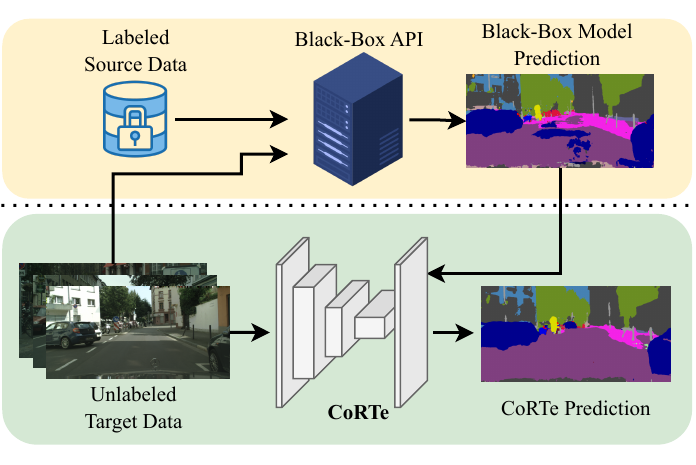} 
  \caption{With {\myalgname} we can train a low-resources model with unlabelled target data extracting knowledge from a pre-trained source model accessible via input-output API. During the knowledge transfer, neither the \textit{source data} nor the \textit{source model} is accessible.} \label{fig:teaser} \vspace{-8pt}
\end{figure}

\section{Introduction}
In the last few years, semantic segmentation models have achieved impressive performances for various applications. The mainstream approach to increase their performance is to design deeper and wider neural networks, promoting accuracy at the expense of computational time, memory consumption, and hardware requirements. However, the continuous development of smart tiny devices, together with the wide diversity of edge applications has underlined the need of delivering models suitable for real-world applications: reduced model footprint and limited inference time. 

Under these regards, a crucial task is to develop methods to transfer large pre-trained models into efficient networks ready for real-world applications. 
However, most of the commercially available architectures are kept as black-box and secured under APIs running on cloud services to minimize model misuse and safeguard against white-box attacks. Furthermore, it is reasonable to expect that the source data used to train the model are confidential or commercially valuable and thus, not released along with the model. 

This task entails two major problems: (i) transferring knowledge from a black-box teacher to an efficient target model, and (ii) addressing the domain gap that exists between the pre-training (source) and the application (target) datasets.  Although the two challenges have been already studied independently, their coupled solution is not straightforward, since most of the model distillation methods assume that teacher and student models share the same data distribution, while unsupervised domain adaptation approaches disregard model efficiency and assume to have access to the source data during the alignment process.

In this paper, we fill this gap and propose a new setting, illustrated in \cref{fig:teaser}, for learning a lightweight semantic segmentation model using an unlabelled target dataset and transferring knowledge from a black-box model trained on a source dataset that  is not provided. 
The black-box model only provides the probabilities associated with the target classes that can be used to supervise the efficient model within the target domain. However, merely optimizing the target network using the predictions generated by the teacher model is vulnerable to inaccuracies that arise as a result of the domain shift between the source and target domains. Specifically, the predictions generated by the source model for the target samples are characterized by a high degree of noise. This limitation strongly vouches for the need of a more sophisticated approach to effectively address the domain shift and enhance the transfer of knowledge between the black-box predictor and the target model.

Motivated by the hypothesis that (i) trained source models yield numerous highly confident yet inaccurate predictions within the target domain and (ii) the efficient model progressively acquires valuable knowledge about the target domain, we propose {\myalgname}, that performs a \textbf{Co}nsistent and \textbf{R}eliable \textbf{T}ransf\textbf{e}r from a black-box source model to train a lightweight semantic segmentation model using unsupervised data. 
Specifically, {\myalgname} extracts reliable pseudo-supervision from the black-box model filtering uncertain pixels based on their relative confidence. Furthermore, to exploit the knowledge learned on the target domain, it refines the pseudo-supervision using the target model. Finally, it trains the student model introducing strong augmentations to improve its generalization abilities.

The contributions of this paper can be summarized as: \vspace{-5pt}
\begin{itemize}[noitemsep]
    \item We study the task of learning a lightweight semantic segmentation model using an unsupervised target dataset and transferring knowledge from a black-box model without being provided with the source dataset.
    \item We propose {\myalgname}, a new method able to extract reliable supervision from the black-box source model and refine it using the knowledge of the target domain.
    \item Comprehensive experiments demonstrate the effectiveness of our proposed method on two challenging synthetic-to-real semantic segmentation protocols.   
\end{itemize} 


\begin{figure*}
\centering
  \includegraphics[width=0.95\textwidth]{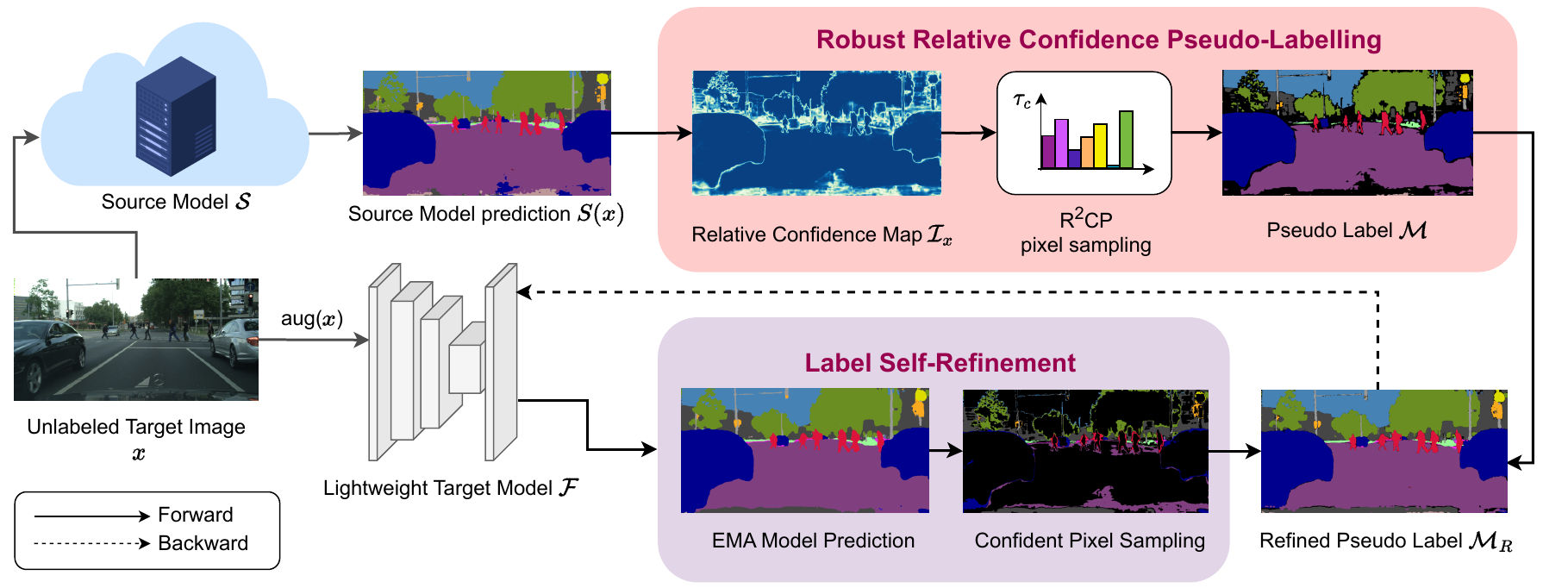}
  \caption{An overview of the proposed {\myalgname} framework. Our approach involves using a teacher model $\mathcal{S}$ to predict labels for an unlabelled target image $x$. To filter the predictions, we introduce a \textbf{Robust Relative Confidence Pseudo-Labelling} method that preserves pixels where the \textit{relative confidence} of the model is above a threshold. The resulting pseudo label $\mathcal{M}$ is then further refined using the \textbf{Label Self-Refinement} technique, which leverages the knowledge gained by the student lightweight model. Finally, the refined pseudo label $\mathcal{M_R}$ is used as the ground truth for training. \vspace{-12pt}} \label{fig:method}
\end{figure*}

\section{Related works}
\myparagraph{Semantic Segmentation.}
It aims to classify each image pixel with its category label.
FCN \cite{long2015fully} was the first to efficiently learn to make dense predictions by replacing the fully connected layers with convolutional layers. This approach evolved in the encoder-decoder architecture \cite{badrinarayanan2017segnet, noh2015learning, ronneberger2015u}. To overcome low spatial resolution at the output, solutions such as dilated convolutions \cite{chen2017deeplab,chen2018encoder,yu2015multi} and skip connections \cite{ronneberger2015u} were proposed. Further improvements have been achieved by harnessing context information \cite{fu2019dual, huang2019interlaced, zhang2018context, chen2017deeplab, zhao2017pyramid, chen2018encoder, poudel2018contextnet}.
Recently, with the growing popularity of attention-based Transformers, they have been effectively adapted to semantic segmentation \cite{zheng2021rethinking,xie2021segmenting, chen2021transunet, liu2021swin}, where they have shown promise in capturing long-range dependencies and context information in images. Motivated by the assumption that traditional frameworks are often computationally demanding and complicated, \cite{NEURIPS2021_64f1f27b} designed an efficient framework that combines Transformers with lightweight multilayer perceptron decoder, and scale the approach to obtain a range of models with varying levels of complexity. \cite{Hoyer_2022_CVPR} extends \cite{NEURIPS2021_64f1f27b} by leveraging the context across features from different encoder levels in order to provide additional information in the decoder, increasing decoder complexity in favor of performance.

\myparagraph{Unsupervised Domain Adaptation (UDA).}
It is a form of Transfer Learning that uses labelled source data to execute new tasks in an unlabelled target domain. In adversarial learning, a discriminator is introduced to reach domain invariance by acting as source-target domain classifier at either intermediate feature level \cite{chen2018road} or output level \cite{vu2019advent,luo2019taking,biasetton2019unsupervised,tsai2019domain}.
While generative-based approaches aim at learning a function to map images across domains  \cite{yang2020fda, zhu2017unpaired, Li_2019_CVPR}, other methods minimize the entropy to force the over-confident source behavior on the target domain \cite{vu2019advent, yang2020fda} or use a curriculum learning approach to gradually infer useful properties about the target domain \cite{zhang2017curriculum, Lian_2019_ICCV}.
Self-training strategies leverage confident predictions inferred from the target data to reinforce the training. To regularize the training, approaches such as confidence thresholding \cite{Zou_2019_ICCV,zou2018unsupervised},  pseudo-label prototypes \cite{zhang2021prototypical, zhang2019category}, prediction ensembling \cite{choi2019self}, consistency regularization \cite{tranheden2021dacs, melas2021pixmatch, prabhu2022augmentation} and multi-resolution input fusion \cite{hoyer2022hrda} have been proposed.

\myparagraph{Source-Free DA.}
Traditional UDA strategies assume the source data is available during training. The concept of \textit{source-free} was introduced by
\cite{chidlovskii2016domain}, motivated by the belief that source data are often subject to commercial or confidentiality constraints between data owners and customers. In the last few years, due to the mounting concerns for data privacy, the source-free setting for UDA is receiving increasing attention.
Recent works \cite{liang2020we, kim2021domain, yang2020unsupervised} fine-tune the model trained on the source domain with the unlabelled target data. Specifically, \cite{liang2020we} adapt the source model with pseudo-labelling and information maximization, which is extended to multi-source in \cite{Ahmed_2021_CVPR, feng2021kd3a}. Other works leverage the source knowledge of the network to synthesize target-style samples \cite{li2020model} or source-like samples \cite{kurmi2021domain, liu2021source} based on the statistics learned by the source model.
In the methods above, the details of the source model are exposed (e.g. \textit{white-box source model}). However, the exposition of the trained source model may be subject to  white-box attacks \cite{Liang_2022_CVPR}. Also, the white-box source model could be unavailable for commercial or safety reasons \cite{zhang2021unsupervised}.
The task of \textit{black-box unsupervised domain adaptation} \cite{zhang2021unsupervised}, where the trained model is fixed and accessible only through an API, has been investigated mostly in image classification. \cite{9512429} divides the target data into two portions to perform self-supervised learning on the uncertain split, while \cite{zhang2021unsupervised} proposes iterative learning with the noisy labels obtained from the black-box model. \cite{Liang_2022_CVPR} reduces the noise through label smoothing during distillation and fine-tunes the model on the target domain.
Differently from these works, we are the first to investigate the challenges of transferring the knowledge of a  black-box to a lightweight semantic segmentation model using only an unsupervised target dataset.

\section{Methodology}

\subsection{Problem Definition}
The goal of \setting\ is to learn a lightweight semantic segmentation model $\mathcal{F}_{\theta}$ using only an unlabelled target dataset $\mathcal{D_T}$. To effectively learn from it, it is possible to exploit a black-box model $\mathcal{S}$ that is accessible only through an API, that has been trained to perform semantic segmentation on a similar yet different source domain.
Formally, it is given a source model $\mathcal{S}: \mathcal{X_S} \xrightarrow{} [0,1]^{|\mathcal{Y}| \times H \times W}$, where $\mathcal{S}$ is fixed and only the per-pixel class probabilities are accessible, $\mathcal{X_S}$ is the source domain image space, and $\mathcal{Y}$ the label space, with $|\mathcal{Y}|$ the number of classes.
In addition, it is provided a target dataset containing $n_t$ unlabelled images from the target domain $\mathcal{D_T} = \{x | x \in \mathcal{X_T}\}$, where $\mathcal{X_T}$ is the target domain image space. Our goal is to train a lightweight segmentation target model $\mathcal{F}_{\theta}: \mathcal{X_T} \xrightarrow{} [0,1]^{|\mathcal{Y}| \times H \times W}$ to infer pixel-wise labels $\{y\}_{i=1}^{H \cdot W}$, where $H, W$ are respectively the height and width of the images, exploiting the dataset $\mathcal{D_T}$ and the supervision coming from the source model $\mathcal{S}$.

\subsection{{Extracting Supervision in the Target Domain}}  \label{sec:pseudo-labeling}

A natural solution to transfer knowledge from a pre-trained larger model to a smaller one is Knowledge Distillation (KD) \cite{44873}, where a small student network is forced to mimic the teacher's predictions. A common technique employed by previous works \cite{Liang_2022_CVPR, UDABlackBox} is to penalize the Kullback-Leibler divergence between the output predictions of the target model (the \textit{student}) and the source model (the \textit{teacher}) on the target domain. 
However, Kim \etal\cite{ijcai2021p362} argue that when the teacher is trained on a dataset with noisy labels, it may transfer corrupted knowledge to the student. We posit that employing KD in our setting would cause the same issue due to the domain shift. Hence, as suggested by \cite{ijcai2021p362}, we focus on \textit{label matching} instead of \textit{logit matching}, forcing the student to neglect the noisy information by only relying on the hard pseudo label produced by the teacher.

However, forcing the student model to perfectly fit the source predictions may lead to reproducing the same inaccuracies of the source model on the target domain, limiting its performance. To effectively address the domain shift between the source and target domain, we argue that it is essential to filter the noisy information coming from the source model by retaining only the reliable pixels to supervise the target model. 
In addition, the pseudo label may be further refined by exploiting the new knowledge learned by the student model on the target domain. 

In the following, we show how to obtain a reliable pseudo label to supervise the student model. First, we illustrate how to filter the noise when extrapolating the pseudo labels from the source model. Thereafter, we refine the pseudo labels directly using the knowledge of the target domain of the model itself.
An illustration of the method is provided in \cref{fig:method}.

\begin{figure}
\centering
\includegraphics[width=0.50\textwidth]
{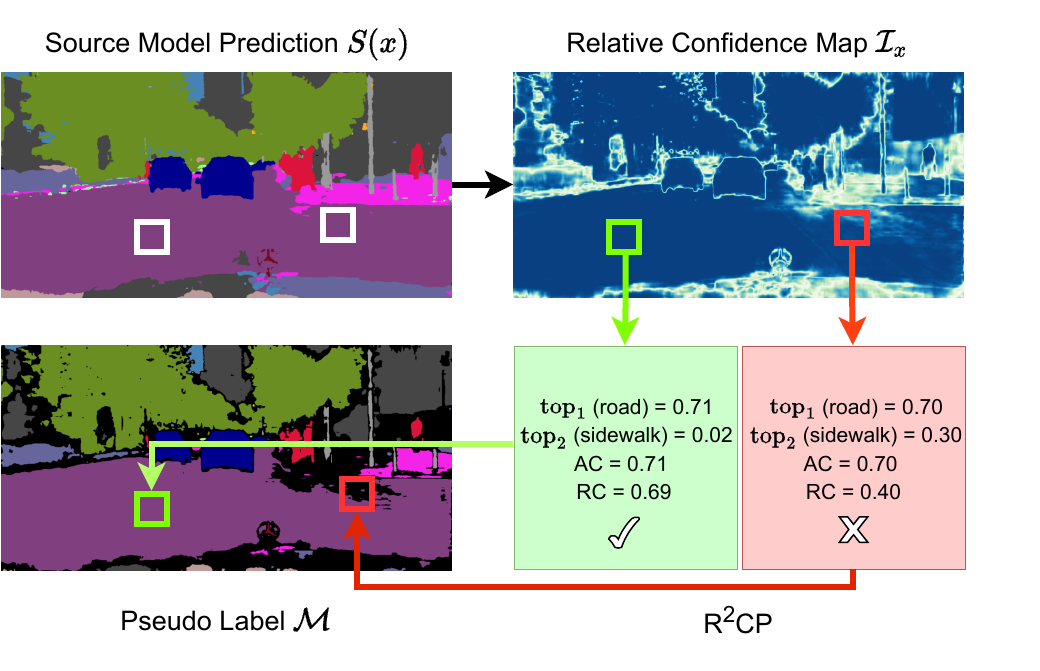}
\vspace{-8pt}
  \caption{In this example, we demonstrate the Robust Relative Confidence Pseudo-Labelling strategy (R$^2$CP) which begins by extracting the Relative Confidence Map $\mathcal{I}_x$ from the source model's prediction on the target image $\mathcal{S}(x)$. The Relative Confidence (RC) is computed using \cref{eq:relative_conf} for each pixel. Finally, we apply \cref{eq:rrcp} to identify the set of reliable pixels to be retained in the Pseudo Label $\mathcal{M}$.
  }
\label{fig:RC} \vspace{-8pt}
\end{figure}

\myparagraph{{Robust Relative Confidence Pseudo-Labelling.}}
To determine the pixels that are favorable for domain transfer, a trivial technique involves applying a pixel-level filter to the pseudo labels on the basis of the model \textit{confidence}, named \textit{absolute confidence} (AC), which serves as an indicator of the network's reliability on each prediction. However, in the case of a noisy teacher model, this approach can produce several confident but incorrect predictions, 
resulting in an unreliable filtering process. This phenomenon is due to the presence of visually distinct categories in the source domain that are more difficult to distinguish in the target domain, making the teacher highly uncertain between the top two predicted classes (i.e. indecision between \textit{bicycle} and \textit{rider} or \textit{sidewalk} and \textit{road}).

A better way to consider the source model confidence is to relate the probability of the predicted class with the other classes. In particular, we consider the \textit{relative confidence} (RC) as the difference between the probability associated with the top-first predicted class and the one associated with the top-second predicted class. Formally, given a target image $x$, we first get the source model probabilities $q = \mathcal{S}(x)$, and we then compute the relative confidence $\mathcal{I}_x$ as:
\begin{equation} 
\label{eq:relative_conf}
    \mathcal{I}^i_{x} = \text{top}_1(q^{i}) - \text{top}_2(q^{i}),
\end{equation}
where $\text{top}_1(\cdot)$ and $\text{top}_2(\cdot)$ indicate the probability value of the first and second predicted classes and $q^{i}$ is the probability distribution for the $i$-th pixel.
Intuitively, the \textit{relative confidence} is a measure of the certainty of the prediction of the teacher network. When $\text{top}_1$ prevails over $\text{top}_2$ the difference is high, meaning that the teacher is certain about the assigned category for the given target pixel $i$. 

Therefore, given a target image $x$, we obtain a reliable pseudo label $\mathcal{M}$ through the \emph{robust RC-driven Pseudo-Labelling} (R$^2$CP) function:
\begin{equation} \label{eq:rrcp}
\mathcal{M}(x)^i_c=
\left\{
\begin{array}{rl}
1 & \text{if}\;  I^i_{x} \ge \tau_{c} \ \text{and} \\  
   & \;\; c = \argmax_{k \in \mathcal{Y}}  q^i_k, \\
0 & \text{otherwise}, \\
\end{array}
\right.
\end{equation} 
where $q^i_k$ is the probability of the pixel $i$ for class $k$, $\mathcal{M}(x)^i_c$ indicates the value of $\mathcal{M}$ for the pixel $i$ and class $c$.
The threshold $\tau_c$ is the average relative confidence $\mathcal{I}$ of the teacher for each class $c$ on the whole target domain:
\begin{equation} 
\tau_c = \frac{1}{N_T^c} \sum_{x \in \mathcal{D}_t}\sum_{i=1}^{H\cdot W} \mathbb{1}(c = \argmax_{k \in \mathcal{Y}} q^i_k) \mathcal{I}^i_{x}, 
\end{equation}
where $N_T^c$ is the number of target samples predicted as $c$ and $\mathbb{1}(c = \argmax_{k \in \mathcal{Y}} q^i_k)$ is the indicator function that equals $1$ when the model predicts the class $c$ for the pixel $i$ and $0$ otherwise.
\cref{fig:RC} illustrates how the Robust Relative Confidence Pseudo-Labelling method works, providing an example of its operation.

\myparagraph{Label Self-Refinement.}
With respect to the source model, which is static, the target model dynamically evolves as training proceeds and gradually learns valuable knowledge about the target domain.
Inspired by \cite{9711265}, we propose to use the target-aware predictions of the target network to refine the pseudo labels, forcing the student to become the teacher model itself over the pixels over which the source model is more uncertain.

In particular, during training, we add the supervision of a second teacher, named $\mathcal{F}_\Theta$, with $\Theta$ indicating its parameters, which is obtained as the temporal ensemble derived via exponential moving average (EMA) \cite{NIPS2017_68053af2} of the target network $\mathcal{F}_{\theta}$. The EMA model is updated based on $\mathcal{F}_{\theta}$ during training following:
\begin{equation} 
\label{eq:EMA}
\Theta_{t+1} = \alpha \Theta_t + (1-\alpha){\theta},
\end{equation}
where $\alpha$ is a parameter controlling the update momentum, $\Theta_t$ e $\Theta_{t+1}$ are, respectively, the weight of the EMA model network before and after the update at the timestep $t$, and we recall $\theta$ are the target model parameters. 

This secondary teacher is used to refine the pseudo labels generated in \cref{eq:rrcp} by including the valuable knowledge provided on the target domain. Our goal is to refine the supervision provided by the source model by introducing a pseudo-supervision on the uncertain pixels by exploiting the confident pixels extracted from the EMA model. 
Formally, given an image $x$ and denoting the predictions of the EMA model as $\hat{p} = {\mathcal{F}_{\Theta}}(x)$, we refine $\mathcal{M}$ as:
\begin{equation} 
\label{eq:mask2}
\mathcal{M}_{R}(x)^i_c=
\left\{
\begin{array}{ll}
   1 & \text{if}\; \mathcal{M}(x)^i_c = 1, \\
   \lambda_t\  &  \text{if}\; \hat{p}^i_c \ge \beta , \mathcal{M}(x)^i_c = 0, \\
    0 & \text{otherwise}, \\
\end{array}
\right.
\end{equation}

where $\beta$ is a confidence threshold applied to the EMA model's probabilities, $\mathcal{M}_{R}(x)^i_c$ and $\hat{p}^i_c$ indicate respectively the refined mask and the probability value of the EMA model at pixel $i$ for class $c$, $\lambda_t$ is a hyper-parameter that controls the contribution of the loss during the training. In particular, $\lambda_t$ linearly increases during the training as the reliability of ${\mathcal{F}_\Theta}$ increases. 

\subsection{Consistent Training of the Target Model}  
The refined pseudo labels $\mathcal{M}_{R}$ provide direct supervision and enable the knowledge transfer between source and target models on the unlabelled target domain. However, the limited size of the target dataset and the label-matching objective between source and target models may have an impact on the generalization capability of the target network.
Several works \cite{berthelot2019mixmatch, french2019semi, xie2020unsupervised, sohn2020fixmatch, CrDoCo, Tavera_2022_CVPR, himix} leverage consistency regularization to make predictions on unlabelled samples invariant to perturbations.
Inspired by these works, we propose to improve the generalization capability of the student by enforcing consistency regularization between the prediction of the teacher model on the original target sample and the prediction of the student on its augmented version.
Specifically, for each training image $x$, we first compute the refined pseudo label $\mathcal{M}_{R}(x)$ as defined in \cref{eq:mask2}. Then, instead of computing the target model's probability on $x$, we augment the image, such that $p(\text{aug}(x)) = \mathcal{F}_\theta(\text{aug}(x))$, where $\text{aug}(\cdot)$ is a function that strongly augments the images without introducing geometric distortions.

Finally, to train the target segmentation model, we use the refined pseudo supervision $\mathcal{M}_{R}(x)$ and the target model's probability obtained on the augmented image $p(\text{aug}(x))$, and we minimize the following loss function:
\begin{equation}
\label{eq:loss hard}
\ell(x) = - \frac{1}{H\cdot W}\sum_{i=1}^{H \cdot W} \sum_{c=1}^C \mathcal{M}_{R}(x)^i_c \log p(\text{aug}(x))^i_c,
\end{equation}
where $p(x)^i_c$ indicates the probability of the target model on the $i$-th pixel and the $c$-th class, and $H$, $W$ are the height and width of the image.
Note that, when $\mathcal{M}_{R}(x)^i_c = 0$ for all $c$, the pixel $i$ does not contribute to the loss function. Differently, if $\mathcal{M}_{R}(x)^i_c \neq 0$, the objective reduces to a weighted cross-entropy loss, where the supervision coming from the source model is weighted $1$, while the supervision coming from the EMA model is weighted $\lambda_t$.

\section{Experiments}
\subsection{Dataset and Evaluation Protocols}
Following \cite{tranheden2021dacs, hoyer2022hrda, yang2020fda, zou2018unsupervised}, we demonstrate the efficacy of the proposed method on the synthetic-to-real unsupervised domain adaptation tasks, where the synthetic source labelled data comes from either GTA5 \cite{10.1007/978-3-319-46475-6_7} or SYNTHIA \cite{Ros_2016_CVPR}, and the unlabelled target data from Cityscapes \cite{Cordts_2016_CVPR}.

\myparagraph{GTA5}: consists of 24,966 training images captured in a video game with resolution 1914$\times$1052. We resize the images to 1280$\times$720 and randomly crop them to 512$\times$512.

\myparagraph{SYNTHIA}: we use the SYNTHIA-RAND-CITYSCAPES subset consisting of 9,400 training images with resolution 1280$\times$760. We randomly crop the images to 512$\times$512.

\myparagraph{Cityscapes}: consists of real-world images  collected from a car in urban environments. We use the 2,975 images from the training set as target data during training.  Previous works resize the training images to 1024$\times$512. To maintain higher resolution, we resize the training images to 1280$\times$640 and randomly crop them to 512$\times$512. 
For a fair comparison, we test on the 500 annotated images from the validation set resized to 1024$\times$512.

We evaluate our method  using the standard segmentation evaluation metrics: classwise Intersection over Union scores (\textbf{IoU}) and mean IoU (\textbf{mIoU}).

\renewcommand{\arraystretch}{1.3}

\begin{table*}[]
\LARGE
    \centering
    \resizebox{0.97\linewidth}{!}{
    \begin{tabular}{l|cc|lllllllllllllllllll|c}
    \hline
    \textbf{Method} & \textbf{SF} & \textbf{T}$\xrightarrow{}$\textbf{S} & 
    \rotatebox{90}{\textbf{Road}} & \rotatebox{90}{\textbf{Sidewalk}} & \rotatebox{90}{\textbf{Building}} & \rotatebox{90}{\textbf{Wall}} & \rotatebox{90}{\textbf{Fence}} & \rotatebox{90}{\textbf{Pole}} &
    \rotatebox{90}{\textbf{T.Light}} & \rotatebox{90}{\textbf{T.sight}} &
    \rotatebox{90}{\textbf{Vegetation}$\ \ \ $} & \rotatebox{90}{\textbf{Terrain}} &
    \rotatebox{90}{\textbf{Sky}}
    &
     \rotatebox{90}{\textbf{Person}} &
      \rotatebox{90}{\textbf{Rider}} &
       \rotatebox{90}{\textbf{Car}}&
        \rotatebox{90}{\textbf{Truck}}&
     \rotatebox{90}{\textbf{Bus}} &
     \rotatebox{90}{\textbf{Train}} &
     \rotatebox{90}{\textbf{Motorbike}} &
     \rotatebox{90}{\textbf{Bicycle}} &
    \textbf{mIoU}\\ 
         \hline  
    
    Source-only & \xmark & \xmark & 78.6 & 22.2 & 80.0 & 34.6 & 27.0 & 36.8 & 45.7 & 27.0 & 86.5 & 37.2 & 84.2 & 67.2 & 36.1 & 80.0 & 47.8 & 50.4 & 42.4 & 38.7 & 25.6 & 49.9 \\
    No adapt & \xmark & \xmark & 4.9 & 12.2 & 62.6 & 9.0 & 10.8 & 11.3 & 12.8 & 8.2 & 83.1 & 28.0 & 61.6 & 44.7 & 1.5 & 54.0 & 30.2 & 14.1 & 0.0 & 4.9 & 1.8 & 24.0 \\ \hline
    DACS\cite{tranheden2021dacs} & \xmark & \xmark  & 85.4 & 0.0 & 83.7 & 5.5 & 4.0 & 26.5 & 25.9 & 40.3 & 85.7 & 39.9 & 87.9 & 45.5 & 0.1 & 80.5 & 28.7 & 21.4 & 0.0 & 0.1 & 1.0 & 34.9\\
    HRDA\cite{hoyer2022hrda} & \xmark & \xmark  & 94.0 & 63.7 & 85.8 & 40.8 & 14.2 & 31.2 & 36.1 & 46.0 & 89.4 & 46.0 & 92.4 & 58.7 & 2.7 & 85.6 & 33.1 & 44.2 & 0.0 & 39.7 & 57.3 & 50.6\\
    \hline
    Naive Transfer &\checkmark & \checkmark  & 
    82.0 &	26.7 &	81.2 &	37.0 &	25.8 &	32.6 &	43.5 &	23.2 &	87.6 &	45.0 &	83.3 &	65.4 &	34.7 &	84.2 &	44.0 &	49.2 &	34.1 &	37.0 &	32.6 &	49.9\\
    KL-DIV &\checkmark &\checkmark & 81.5 &	26.2 &	80.8 &	36.7 &	25.0 &	32.4 &	43.1 &	23.5 &	87.6 &	45.0 &	83.5 &	65.3 &	34.7 &	84.1 &	44.6 &	49.8 &	33.8 &	37.8 &	32.9 &	49.9\\
    \hline
    \textbf{{\myalgname}} &\checkmark &\checkmark  &
    87.0 &	37.5 &	84.6 &	44.6 &	29.0 &	31.0 &	41.5 &	25.4 &	88.0 &	46.4 &	88.3 &	62.2 &	33.9 &	86.4 &	54.2 &	61.6 &	52.0 &	44.2 &	56.5 &	\textbf{55.5} \\ \hline
        Target-only &\checkmark & \xmark & 97.6 &	80.9 &	90.4 &	54.7 &	53.4 &	50.9 &	58.4 &	68.6 &	90.3 &	58.6 &	93.0 &	73.9 &	52.8 &	92.3 &	60.3 &	76.1 &	53.6 &	52.4 &	69.1 &	69.9 \\ \hline
    \end{tabular}
}     \caption{mIoU for GTA5$\xrightarrow{}$Cityscapes. \textbf{SF} denotes the \textit{Source-Free} methods, whereas \textbf{T}$\xrightarrow{}$\textbf{S} refers to the methods that leverage the black-box model for training the target network. All methods use B0 as encoder, while \textit{Source-only} uses B5.}    \label{tab:gta2cs}
\end{table*}

\renewcommand{\arraystretch}{1.3}
\begin{table*}[]
\LARGE
    \centering
    \resizebox{0.97\linewidth}{!}{
    \begin{tabular}{l|cc|llllllllllllllll|c}
    \hline
    \textbf{Method} & \textbf{SF} & \textbf{T}$\xrightarrow{}$\textbf{S} & 
    \rotatebox{90}{\textbf{Road}} & \rotatebox{90}{\textbf{Sidewalk}} & \rotatebox{90}{\textbf{Building}} & \rotatebox{90}{\textbf{Wall}} & \rotatebox{90}{\textbf{Fence}} & \rotatebox{90}{\textbf{Pole}} &
    \rotatebox{90}{\textbf{T.Light}} & \rotatebox{90}{\textbf{T.sight}} &
    \rotatebox{90}{\textbf{Vegetation}$\ \ \ $} &
    \rotatebox{90}{\textbf{Sky}}
    &
     \rotatebox{90}{\textbf{Person}} &
      \rotatebox{90}{\textbf{Rider}} &
       \rotatebox{90}{\textbf{Car}}&
     \rotatebox{90}{\textbf{Bus}} &
     \rotatebox{90}{\textbf{Motorbike}} &
     \rotatebox{90}{\textbf{Bicycle}} &
    \textbf{mIoU}\\ 
         \hline  
    Source-only & \xmark & \xmark  & 58.9 & 22.2 & 79.3 & 22.9 & 1.0 & 40.6 & 34.7 & 21.2 & 81.8 & 80.5 & 58.2 & 20.8 & 78.5 & 28.6 & 16.8 & 20.5 & 41.6 \\ 
    No adapt & \xmark & \xmark  & 12.9 & 14.8 & 55.9 & 3.6 & 0.0 & 20.8 & 3.8 & 6.2 & 66.1 & 63.5 & 46.7 & 8.9 & 37.4 & 5.9 & 1.6 & 9.6 & 22.3 \\ \hline
    DACS\cite{tranheden2021dacs}& \xmark & \xmark  & 69.2 & 14.3 & 72.8 & 3.5 & 0.2 & 32.2 & 7.2 & 29.6 & 84.4 & 83.4 & 58.0 & 12.3 & 78.2 &  0.3 & 0.1 & 10.0 & 34.7 \\
    HRDA\cite{hoyer2022hrda} & \xmark & \xmark & 70.2 & 29.0 & 83.3 & 0.8 & 0.2 & 39.0 & 34.8 & 41.6 & 85.6 & 92.4 & 66.8 & 5.6 & 80.0 & 0.0 & 0.0 & 59.7 & 43.1\\
    \hline
    Naive transfer &\checkmark & \checkmark  &
    65.2 &	25.1 &	80.8 &	10.0	& 1.3	& 40.5 &	36.5	& 19.1 &	83.7 &	83.1 &	57.6 &	20.6	& 80.5 &	40.2 &	19.5 &	31.4 & 44.0\\
    KL-DIV &\checkmark &\checkmark  &
    65.2 &	25.1 &	80.6 &	17.8 &	1.3 &	40.0 &	35.0 &	19.0 &	83.7 &	83.0 &	57.5 &	21.0 &	79.4 &	39.5 &	20.0 &	31.5 &	43.7\\
    \hline
    \textbf{{\myalgname}} &\checkmark &\checkmark  &
    68.4 &	26.7 &	83.1 &	22.9 &	1.7 &	38.7 &	38.3	& 20.3 &	85.8 &	85.4 &	56.7 & 18.9	& 85.0 &	47.9 &	21.7 &	31.3 &	\textbf{45.8} \\ \hline 
    Target-only &\checkmark & \xmark &
    97.6 &	 80.9 &	90.4 &	54.7 &	53.4 &	50.9 &	58.4 &	68.7 &	90.3 &	93.0 &	73.9 & 	52.8 &	92.2 &	60.3 &	52.4 &	69.1 	& 71.2\\ \hline
    \end{tabular}
}
    \caption{mIoU for SYNTHIA$\xrightarrow{}$Cityscapes. \textbf{SF} denotes the \textit{Source-Free} methods, whereas \textbf{T}$\xrightarrow{}$\textbf{S} refers to the methods that leverage the black-box model for training the target network.  All methods use B0 as encoder, while \textit{Source-only} uses B5.}    \label{tab:syn2cs} \vspace{-7pt}
\end{table*}

\subsection{Baselines} Black-box unsupervised domain adaptation for Semantic Segmentation is fairly new. Therefore, we implemented several baselines. 
\textbf{Source only}: evaluates the performance of the trained source model on the target images. \textbf{No adapt}: the target network is trained on the annotated source domain without any adaptation.  \textbf{DACS} \cite{tranheden2021dacs} and \textbf{HRDA} \cite{hoyer2022hrda}: provide adaptation during the target network training. \textbf{Naive transfer}: the target images are pseudo-labelled by the source model and used to train the target network.
\textbf{KL-DIV}: we train the target model by penalizing the KL-divergence between the output predictions of the student
and the teacher. \textbf{Target-only}: the target network is directly trained with the annotated target domain.

\myparagraph{Implementation Details}
We employ the Transformers-based architectures tailored for semantic segmentation proposed in \cite{NEURIPS2021_64f1f27b}.
The source model is based on DAFormer \cite{Hoyer_2022_CVPR}. It consists of a MiT-B5 encoder \cite{NEURIPS2021_64f1f27b} and a context-aware feature fusion decoder \cite{Hoyer_2022_CVPR}. As the target model we employ the lightweight SegFormer-B0 \cite{NEURIPS2021_64f1f27b}.
We pre-train the target network on the ImageNet-1K and randomly initialize the decoder. Following \cite{Hoyer_2022_CVPR}, we train the network with AdamW optimizer using a learning rate of $ 6\times 10^{-5}$ for the encoder and $6\times 10^{-4}$ for the decoder, weight decay of 0.01, and linear learning rate warm up.
We use a batch size of 8 and train the model for 80k iterations. We set $\alpha = 0.99$, $\beta=0.60$, $\lambda $ in the range [0,5]. To enforce consistency regularization, we apply Color jittering, Gaussian blur, and random flipping.
Following previous works \cite{Hoyer_2022_CVPR}, the source network is optimized in a supervised manner by minimizing the cross-entropy loss on the source domain for 40k iterations using batches of 2 images.

\begin{figure*}
     \centering
     \begin{subfigure}[b]{\textwidth}
        \includegraphics[width=\textwidth]{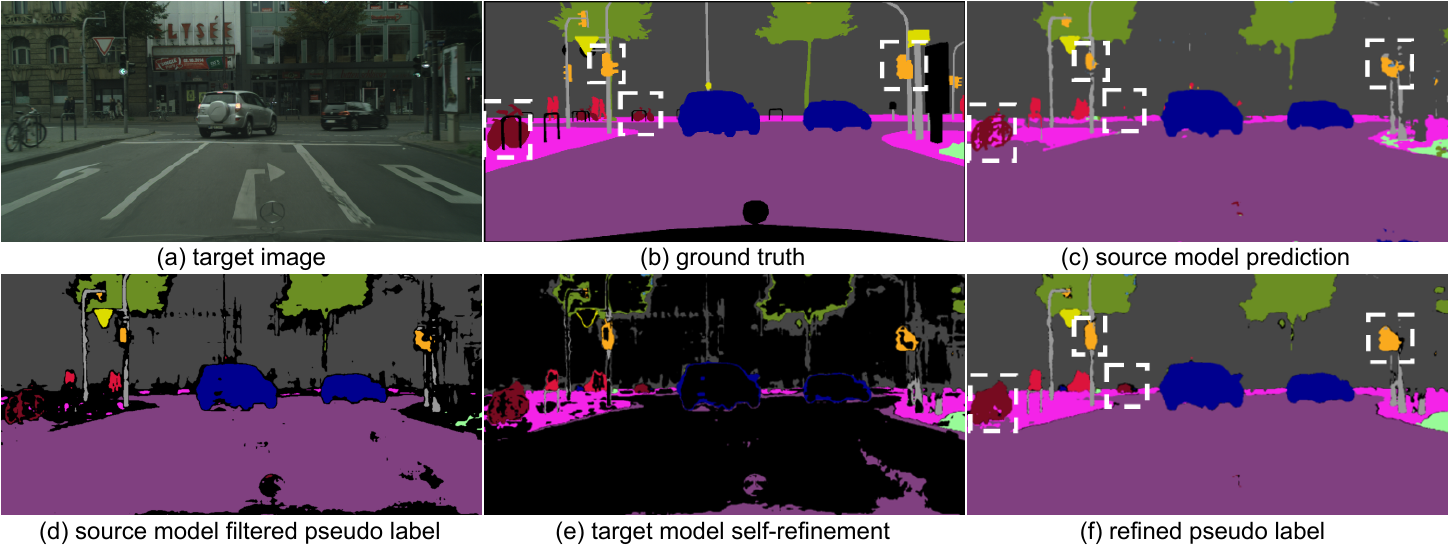}
     \end{subfigure}
     \caption{\textit{Graphical interpretation of the label generation process.} At the top left, a target image from Cityscapes (a) and its corresponding label (b). We query the teacher model and obtain its prediction for the target image (c). Our robust pseudo-labelling module exploits the relative confidence of the teacher model to filter out the uncertain pixels (black in d). During the training process, the increasing knowledge of the student on the target data is used to automatically refine the pseudo label (e) resulting in the final label for training the model (f).
     }
     \label{fig:qualitatives} \vspace{-10pt}
\end{figure*}

\subsection{Results}
\myparagraph{GTA5 $\xrightarrow{}$ Cityscapes} \\
In \cref{tab:gta2cs} we show the results of our proposed framework when the source model is trained on GTA5.
When evaluated on the target domain, the source model achieves an overall mIoU of 49.9\%, performing well on simple classes (\textit{car, sky, road}), but failing on other classes (\textit{sidewalk}, \textit{bicycle}) where the discrepancy between the source and target domains has a significant impact. The domain shift is more pronounced on the lightweight target model, which achieves an overall mIoU of 24\% (\textit{No adapt}) indicating lower generalization capability on the new domain. Moreover, the \textit{Target-only} upper bound demonstrates a high domain shift between GTA5 and Cityscapes: it achieves 69.9\%, which is 45.9\% higher than \textit{No adapt}. Standard UDA techniques improve the model's performance significantly, with DACS \cite{tranheden2021dacs} achieving 34.9\% and HRDA \cite{hoyer2022hrda} achieving 50.6\%. These results show that UDA techniques help in adapting the network, improving its performance. However, they rely on source data, which are not available in our setting. 

When directly trained with the knowledge generated by the source model (\textit{Naive transfer} and \textit{KL-DIV}), the student converges to the performance of the teacher (\textit{Source-only}), obtaining very similar results. The target network achieves satisfying results on the unlabelled target domain (-20\% \wrt \textit{Target-only}), yet it copies the source model behavior, fitting also its noisy predictions. 
{\myalgname} outperforms them by nearly +5.6\%, obtaining a gain in almost all the classes. Significant improvements are observed particularly in the classes over which the teacher is more uncertain, namely \textit{sidewalk} (+11\%) which is typically misclassified as \textit{road}, and \textit{bicycle} (+24\%) which is often confused with \textit{rider}. Additionally, a substantial improvement is obtained for the typically hard-to-transfer class \textit{train} (+18\%), where it almost reaches the \textit{Target-only} upper bound (52\% vs. 53.6\%). 
Compared with \textit{HRDA}\cite{hoyer2022hrda} for UDA, {\myalgname} enables competitive performance (+4.9\%), while training with images at lower resolution (512$\times$512 vs. 1024$\times$1024) and with no access to the annotated source dataset. 

\myparagraph{SYNTHIA $\xrightarrow{}$ Cityscapes} \\
In \cref{tab:syn2cs} we report the results of training the source model on SYNTHIA. Following the protocol of UDA, we report mIoU on the 16 classes in common with Cityscapes.

The results are remarkably coherent with the previous setting. Specifically, the source model achieves an overall mIoU of 41.6\%, yet it suffers from the domain shift on hard classes such as \textit{bicycle} or \textit{motorbike}. In comparison, the B0 model trained on the source data (\textit{No adapt}) achieves largely worse performance (22.3\%), showing the larger model has better generalization capabilities. 
Moreover, the \textit{Target-only} upper bound performance confirms the domain shift between SYNTHIA and Cityscapes: it achieves 71.2\%, which is 48.9\% higher than the \textit{No Adapt} baseline.
Employing standard UDA techniques, the performances improve sensibly: DACS \cite{tranheden2021dacs} achieves 34.7\% and HRDA \cite{hoyer2022hrda} 43.1\%. Differently, methods relying on the knowledge of the source model do not use source images while achieving comparable performance. In particular, \textit{Naive Transfer} and \textit{KL-DIV} achieve results slightly better than HRDA \cite{hoyer2022hrda} (respectively 44\% and 43.7\%).
Finally, we show that {\myalgname} outperforms all the baselines, achieving an overall IoU of 45.8\%. Specifically, it improves HRDA \cite{hoyer2022hrda} of +2.7\%. 
In addition, it outperforms the \textit{Naive transfer} baseline of 1.8\%, showing the benefits of filtering the pseudo labels coming from the source model and refining them using the target model knowledge. 

\begin{figure*}
\centering
  \includegraphics[width=0.95\textwidth]{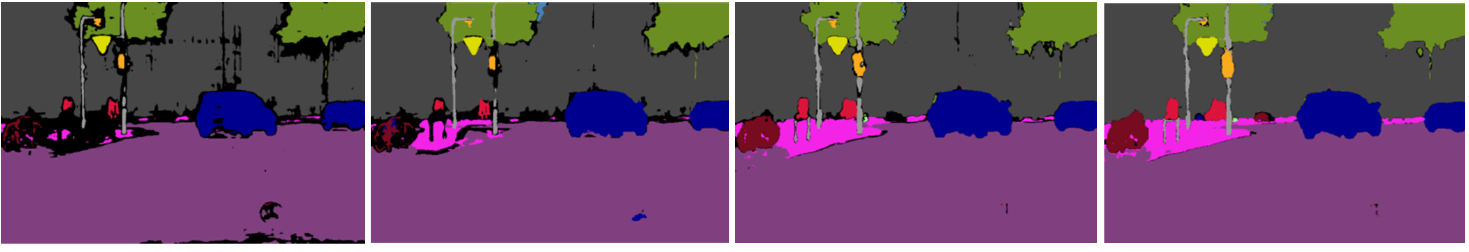}
  \caption{\textit{Self-label refinement}. 
  Visual representation of our refined pseudo label used at different steps of the training (from left to right at 0, 1.5k, 5k, and 80k steps respectively). Intuitively, at the very beginning of the training, the teacher's prediction is filtered from the uncertain pixels ad used to train the student. During training, this latter gradually increases its confidence and its own predictions can be used to refine the pseudo label with our Label Self-Refinement module. 
  }
  \label{fig:ssl} \vspace{-6pt}
\end{figure*}

\begin{figure}
\centering
\includegraphics[width=0.99\columnwidth]{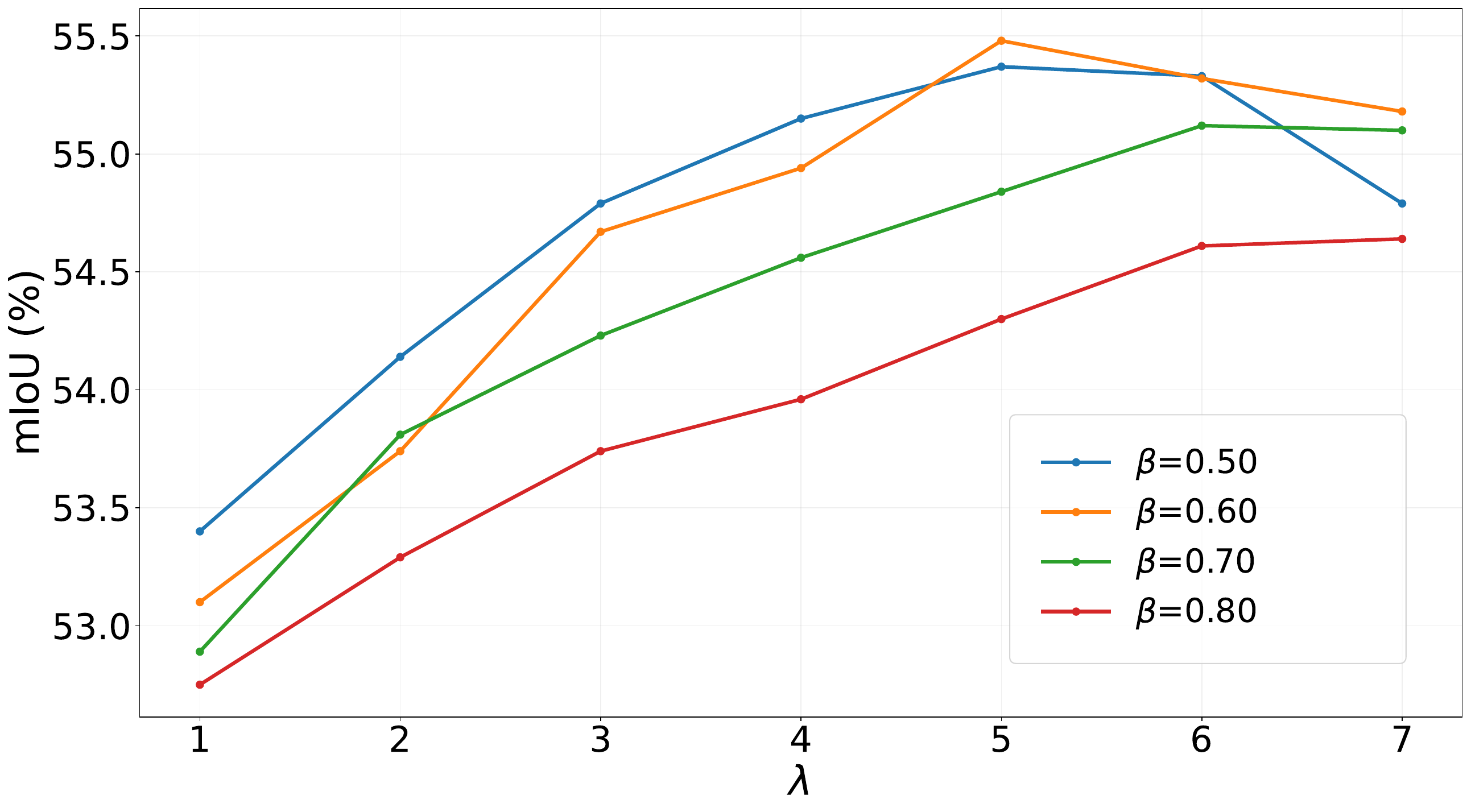}
\vspace{-10pt}
  \caption{Parameters selection for $\lambda_t$ and $\beta$.}
    \label{fig:ablation_parameters} \vspace{-12pt}
\end{figure}

\subsection{Ablation Study}
\myparagraph{Influence of each component.}
In this paragraph, we dissect the contributions of each component to the overall performance. In \cref{tab:ablation}, we report the results when the source model is trained on GTA5. 
We initially show the baseline performance of the target model (line 1) trained under the supervision of the noisy pseudo labels produced by the source model (\textit{Naive Transfer}). The addition of our Robust RC Pseudo-Labelling module (line 2) yields an improvement in terms of performance of 1.5\%. Combined with Consistency Regularization (line 4), the mIoU increases up to 52\%, enabling a further gain of 0.6\%. The most significant contribution, however, is granted by the Label Self-Refinement (line 5), which ensures a further improvement of 3.5\% in the final mIoU. 
In addition, we also evaluate the contribution of our Robust RC Pseudo-Labelling function with respect to the filtering function based on Absolute Confidence (AC Filtering* in \cref{tab:ablation}) (line 3), proving the effectiveness of our certainty-driven filtering approach.

\myparagraph{Parameter Sensitivity Analysis.} Our student-driven label refinement involves two hyperparameters $\beta$ and $\lambda_t$. To investigate their impact on the training process, we conduct experiments on GTA5$\xrightarrow{}$ Cityscapes. In  \cref{fig:ablation_parameters} we report the resulting mIoU while changing $\lambda_t$ and $\beta$. The experimental results demonstrate that our proposed model achieves the highest mIoU with the values of $\lambda_t=5$ and $\beta=0.60$.

The performance of the model is comparatively lower when $\beta$ is set to 0.80 as a substantial number of pixels are filtered out. The model's performance is observed to improve upon decreasing the threshold value, as a greater number of informative pixels are included in the supervision process. 
When increasing the parameter $\lambda_t$, the mIoU rapidly increases until it reaches a plateau within the range [5, 6, 7]. This result proves that increasing the contribution of the target network in the training process significantly enhances the final performance of the network.

\myparagraph{Qualitative Analysis.}
In \cref{fig:qualitatives} we report a qualitative interpretation of the label generation process. A comparison between the refined pseudo label (\cref{fig:qualitatives}f) and the prediction obtained from the source model (\cref{fig:qualitatives}c) reveals the benefit of the self-refinement process, which enables a higher level of detail (e.g. the \textit{traffic light} in the foreground) and even the identification of entire objects (e.g. the \textit{bicycle} in the background).
The process of self-refinement gradually includes valuable knowledge during training, as shown in \cref{fig:ssl}, where progressively larger portions of images are added to the supervision. 
\renewcommand{\arraystretch}{1.0}
\begin{table}[]
    \small
    \centering
    \resizebox{1.0\linewidth}{!}{
    \begin{tabular}{ccccc}
    \toprule
     \thead{R$^2$CP}  & \thead{AC \\Filtering*} &
    \thead{Consistency \\ Regularization} &
    \thead{Label self- \\ Refinement} & \thead{mIoU} \\
    \midrule
    \xmark & \xmark & \xmark & \xmark  & 49.9 \\
    \midrule
    \checkmark & \xmark & \xmark & \xmark & 51.4\\
 \xmark & \checkmark & \xmark &  \xmark & 50.5\\
     \midrule\checkmark & \xmark & \checkmark & \xmark  & 52.0 \\
     \midrule
     \checkmark & \xmark & \checkmark & \checkmark& \textbf{55.5}\\
     \bottomrule     
     
    \end{tabular}
}   \caption{Ablation study on GTA5$\xrightarrow{}$Cityscapes.} \vspace{-5pt}
    \label{tab:ablation}
\end{table}

\section{Conclusion}
In this paper, we explore the challenging scenario of learning a compact and efficient neural network for semantic segmentation by leveraging a black-box model without access to any source data or target annotations. 
To address this novel setting, we propose {\myalgname} that reliably transfers the knowledge from the black-box predictor and provides valuable pseudo-supervision from the target model itself during training. 
We assess the benefits of {\myalgname} on two synthetic-to-real benchmarks, showing it is able to outperform all the considered transfer learning baselines.

\myparagraph{Limitations}
{\myalgname} enables efficient knowledge transfer between a black-box source predictor and a lightweight target model, allowing it to operate on unlabelled target domains. However, it has limitations in dealing  with unknown classes present in the target domain that were not learned by the source model. Additionally, to reach state-of-the-art results, {\myalgname} requires a robust pre-trained source model.   

\small{
\myparagraph{Acknowledgements}
This study was carried out within the FAIR - Future Artificial Intelligence Research and received funding from the European Union Next-GenerationEU (PIANO NAZIONALE DI RIPRESA E RESILIENZA (PNRR) – MISSIONE 4 COMPONENTE 2, INVESTIMENTO 1.3 – D.D. 1555 11/10/2022, PE00000013). This manuscript reflects only the authors’ views and opinions, neither the European Union nor the European Commission can be considered responsible for them.
}


{\small
\bibliographystyle{ieee_fullname}
\bibliography{egpaper_for_review}
}

\end{document}